\documentclass[10pt,twocolumn,letterpaper]{article}

\usepackage{cvpr}
\usepackage{times}
\usepackage{epsfig}
\usepackage{graphicx}
\usepackage{amsmath}
\usepackage{amssymb}
\usepackage{xcolor}
\usepackage{makecell}

\usepackage{subcaption}

\usepackage[pagebackref=true,breaklinks=true,letterpaper=true,colorlinks,bookmarks=false]{hyperref}

\cvprfinalcopy 

\def\cvprPaperID{1676} 

\newcommand{\comment}[1]{}

\ifcvprfinal\fi
\begin{document}

\title{Using Scene Graph Context to Improve Image Generation}


\author{Subarna Tripathi\\
Intel AI Lab\\
\and
Anahita Bhiwandiwalla\\
Intel AI Lab\\
\and
Alexei Bastidas\\
Intel AI Lab\\
\and
Hanlin Tang\\
Intel AI Lab\\
{\tt\small \{subarna.tripathi, anahita.bhiwandiwalla, alexei.bastidas\}@intel.com}
}

\maketitle

\label{abs}
\begin{abstract}

Generating realistic images from scene graphs asks neural networks to be able to reason about object relationships and compositionality. As a relatively new task, how to properly ensure the generated images comply with scene graphs or how to measure task performance remains an open question. In this paper, we propose to harness scene graph context to improve image generation from scene graphs. We introduce a scene graph context network that pools features generated by a graph convolutional neural network that are then provided to both the image generation network and the adversarial loss. With the context network, our model is trained to not only generate realistic looking images, but also to better preserve non-spatial object relationships. We also define two novel evaluation metrics, the relation score and the mean opinion relation score, for this task that directly evaluate scene graph compliance. We use both quantitative and qualitative studies to demonstrate that our proposed model outperforms the state-of-the-art on this challenging task.
\end{abstract}

\section{Introduction}
\label{sec:intro}
The generation of realistic scenes marks an important challenge for neural networks, with recent advancements enabling synthesizing high-resolution images, even when they are conditioned on class labels\cite{cgan}, captions \cite{ma_reed16}, or latent dimensions \cite{prog-gan}. However, the ability to interpret object sizes, relationships, and composition to synthesize realistic scenes still eludes neural networks. For example, state-of-the-art methods for caption-conditioned image generation still struggle to generate realistic images across a broad vocabulary.

Johnson \etal \cite{img_gen_from_scene_graph18} recently explored, instead, to generate images from scene graphs. Compared to captions, scene graphs are a more structured representation, with objects as nodes, and edges marking the semantic relationship between objects. This method yields significantly improved generated images, but notably struggles with cluttered or small objects. 

\begin{figure}[!t] \begin{center}
\begin{subfigure}[b]
{0.48\textwidth}
    \includegraphics[width=\textwidth]{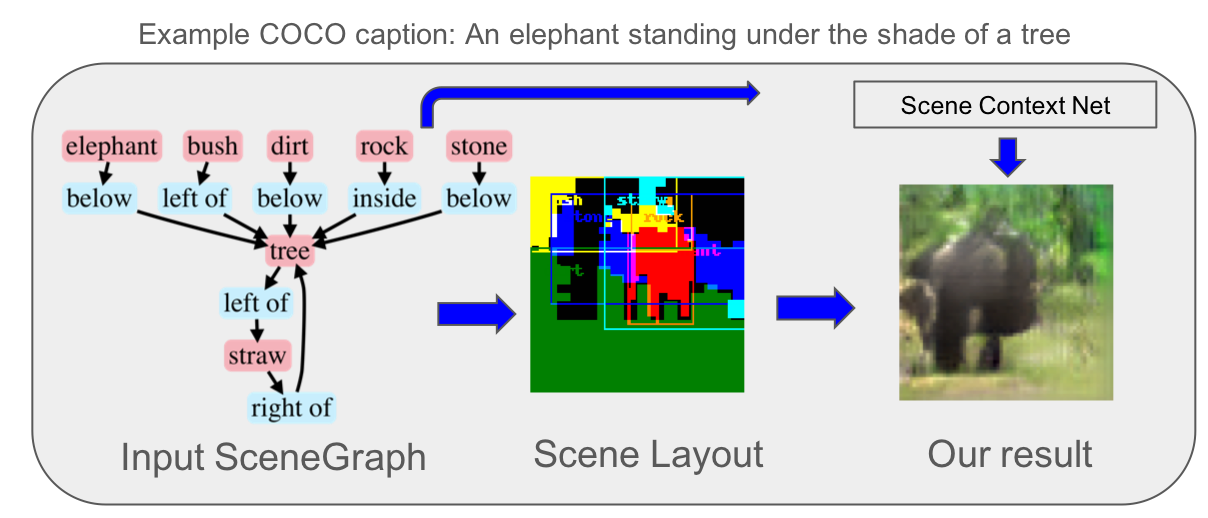}
    \end{subfigure}
 \begin{subfigure}[b]{0.16\textwidth}
    \includegraphics[width=\textwidth]{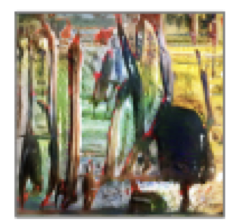}
    \caption{StackGAN \cite{stackGan17}}
 \end{subfigure}
 \begin{subfigure}[b]
 {0.15\textwidth}
    \includegraphics[width=\textwidth]{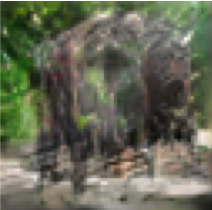}
    \caption{Johnson \etal \cite{img_gen_from_scene_graph18}}
 \end{subfigure}
 \begin{subfigure}[b]
 {0.15\textwidth}
    \includegraphics[width=\textwidth]{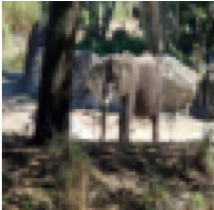}
    \caption{Real Image}
 \end{subfigure}
\end{center}
\caption{Overall framework of the proposed method. Given a scene graph, our algorithm constructs a semantic layout of the scene. With the semantic layout and the scene context network, the model generates an image conditioned on the inferred layout and scene context. Compare our result against previous work in (a-b). Best viewed in color.}
\label{fig:intro}
\end{figure}

In this paper, we improve upon the work in \cite{img_gen_from_scene_graph18} in several ways. First, we introduce a scene context network to provide context features to both the generator and discriminator, which incentives compliance of the generated images to the scene graph. Second, we borrow context-aware loss from the text-based methods to provide an additional image-graph matching signal. Lastly, generating images from scene graph is a relatively new task without well-defined metrics. We introduce novel evaluate metrics more suited for this task: the relation score and mean opinion relation score (MORS), both of which measures the compliance of generated images to the scene graph.

Based on our experiments on the Visual Genome\cite{visual_genome16} and COCO-stuff \cite{coco-stuff} datasets, our proposed model (Figure \ref{fig:intro}) establishes a new state-of-the-art on this tasks, outperforming the Johnson \etal \cite{img_gen_from_scene_graph18} model on both quantitative and qualitative scores. For Visual Genome, which includes semantic relationships, our model ensures better compliance with those semantic relationships, as measured with our mean opinion relation score (MORS).

\section{Related Work}
\label{sec:related}

Scene graphs provide a structured description of complex scenes \cite{battaglia2018relational} and the semantic relationship between objects. Generating scene graphs from images is relatively well studied \cite{neural_motifs_2017, xu2017scenegraph, jnt_embedding17, GIP2018}. Proposed approaches are remarkable in their diversity, include augmenting recurrent neural networks with message passing \cite{xu2017scenegraph}, or repurposing models from keypoints detection \cite{hourglass_NewellYD16} to detect objects and edges with associate embeddings \cite{pixel_to_graph17}. Zellers et al \cite{neural_motifs_2017} find that certain subgraphs (motifs) appear regularly in scene graphs, and that object categories are strong predictors of likely relationships. They exploit these findings to introduce building global context with recurrent neural networks. Scene graphs have also been explored for image retrieval tasks \cite{jnt_embedding17,SG_for_IR_15}.

Image generation from scene graphs, however, are relatively new. Johnson \etal \cite{img_gen_from_scene_graph18} extract objects and features from a scene graph with a graph convolutional neural network. A network then applies these features to predict a scene layout of objects, which are then used by a cascade refinement network \cite{crn} to generate realistic images. The closest other image generation methods are usually conditioned on text. Text to image generation has a rich prior-art. Recently most promising methods \cite{Reed_text_to_image_2016, stackGan17, reed_draw_16} are those that are based on conditional Generative Adversarial Networks (GAN).

\section{Method}
\label{sec:method}
\begin{figure*}[!t] \begin{center}
\begin{subfigure}[b]
{0.9\textwidth}
    \includegraphics[width=\textwidth]{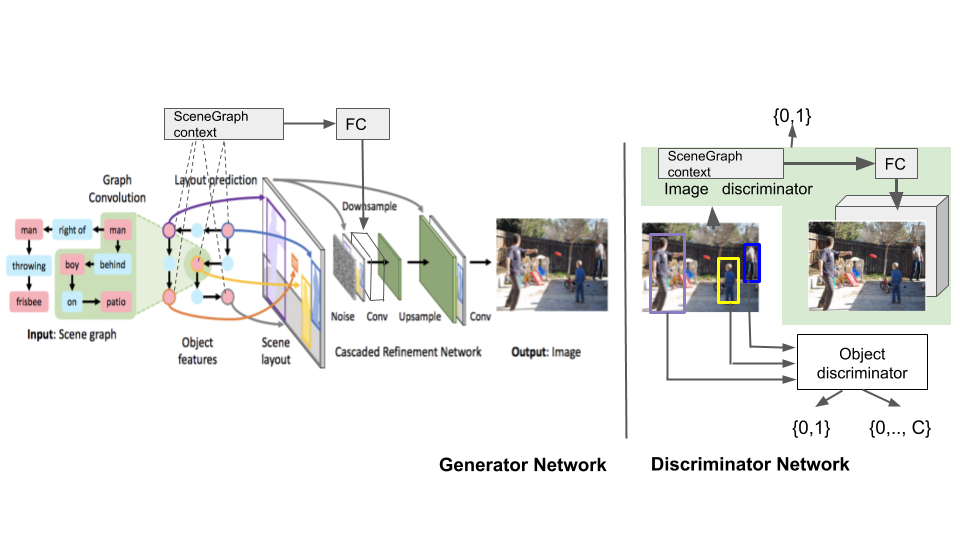}
  \end{subfigure}
  %
\caption{Overview of the proposed image generation network. Image generation is conditioned on the Scene Graph context and the semantic layout generated by the graph convolutional neural network. It generates an image that matches both the inputs, scene layout and aggregated scene context. The image discriminator is also conditioned on the scene-graph context. FC denotes fully-connected layer.}  
\label{fig:pipeline}
\end{center}\end{figure*}

The overall pipeline of the image generation framework is illustrated in figure \ref{fig:pipeline}. Given a scene graph consisting of objects and their relationships, our model constructs realistic image corresponding to the scene graph. Our framework is built upon \cite{img_gen_from_scene_graph18}. Briefly, the scene graph is converted into object embedding vectors from a Graph Convolution Neural Network (GCNN), which are then used to predict bounding boxes and segmentation masks for each object. These are combined to form a scene layout as an intermediate between the graph and the image domains. Finally, a Cascade Refinement Network \cite{crn} generates the image. We improved upon this framework in several ways, which we introduce below.

\paragraph{Scene graph context:} In the original formulation, the adversarial loss only encourages the image patches to appear realistic, but not necessarily comply with the scene graph's object relationships. In our model, we add a scene graph context network that pools the features generated from the graph convolutional neural network. These pooled context features are then passed to a fully-connected layer that generates embeddings that are provided to both the generator and the discriminator networks during training. The scene context network encourages the images not only to appear realistic, but to respect the scene graph relationships.

In our scene context network, the output dimension of the fully connected layer in the generator is $8$. The value for the same in the discriminator network is $4$. The image discriminator of the scene context network can optionally use the layout as additional input. However, we found using the layout in the image  discriminator in this end-to-end framework often leads to mode collapse. 




\paragraph{Matching-aware Loss:} To further encourage the model to generate images that match input scene graph descriptions, we employ a matching-aware loss. Matching-aware loss have been used for matching input text descriptions in the literature \cite{ma_reed16, Hong2018CaptionLayout}. We denote a ground-truth training example as $(M, s, X)$, where $M$, $s$ and $X$ denote semantic layout, scene graph embedding and image, respectively. We then construct an additional mismatching triplet $(M, \hat{s}, X)$ by sampling random scene graph embedding $\hat{s}$ non-relevant to the image. We add these mismatching triplets as additional fake examples in adversarial training, and extend the conditional adversarial loss for image generator.

The image generator $G_{img}$ is conditioned on both the inputs, the scene layout $M$ and the aggregated scene context embedding $s$. It is jointly trained with the discriminators $D_{img}$ and $D_{obj}$. The generation network is trained to minimize the weighted sum of six losses \cite{img_gen_from_scene_graph18}, now with the matching aware loss:

\begin{itemize}
    \item Bounding box loss, $L_{box}$ penalizing the $L_1$ difference between the ground truth and predicted bounding box co-ordinates
    
    \item Mask loss, $L_{mask}$, penalizing differences between the ground truth and predicted masks with pixel-wise cross entropy; used only for COCO-stuff where ground truth mask is available
    
    \item Pixel reconstruction loss $L_{pix}$ that penalizes $L_1$ differences between the RBG of ground truth and generated images
    
    \item Adversarial image loss ${L_{GAN}}^{img}$ from $D_{img}$ employing \emph{matching-aware loss} that encourages images to be both realistic and relevant to scene context
    
    \item Adversarial object loss ${L_{GAN}}^{obj}$ from $D_{obj}$ encouraging objects to appear realistic
    
    \item Auxiliary classifier loss ${L_{AC}}^{obj}$ from $D_{obj}$ encouraging each generated object be classified by the object discriminator
    
\end{itemize}


\paragraph{Implementation Details:} We used the same augmentation scheme and graph convolutional network as \cite{img_gen_from_scene_graph18}. Scene Graph context pooling uses \emph{sum} pooling, which performs better than \emph{average} pooling for this scene context network. We used Adam \cite{adam_KingmaB14} optimizer and a batch size of $32$ for our experiments. Training a single model took about 5 days on one NVIDIA Pascal GPU.


\section{Experiments}
\label{sec:results}
We train our model to generate $64\times 64$ images on the
Visual Genome \cite{visual_genome16} and COCO-Stuff\cite{coco-stuff} datasets. 
In our
experiments we aim to show that the generated images look realistic and they
respect the objects and relationships
of the input scene graph.

\subsection{Datasets}

\textbf{COCO}: We performed experiments on the 2017 COCO-Stuff
dataset \cite{coco-stuff}, which augments a subset of the COCO
dataset \cite{coco} with additional stuff categories. The dataset annotates
$40K$ train and $5K$ val images with bounding boxes
and segmentation masks for $80$ thing categories (people,
cars, etc.) and $91$ stuff categories (sky, grass, etc.).
Similar to \cite{img_gen_from_scene_graph18},  we used thing and stuff annotations to construct synthetic scene
graphs based on the 2D image coordinates of the objects,
encoding six mutually exclusive geometric relationships: `left'
`of', `right of', `above', `below', `inside', and `surrounding'.
We ignored objects covering less than 2\% of the image,
and used images with $3$ to $8$ objects.

\textbf{Visual Genome}: We experimented on Visual Genome [26]
version 1.4 (VG) which comprises 108,077 images annotated
with scene graphs. 
Similar to the pre-processing described in \cite{img_gen_from_scene_graph18}, 
we used object and relationship categories
occurring at least 2,000 and 5,00 times, respectively in
the training set. The resulting training set included 178 object and 45 relationship types.
We ignored small objects, and only selected images with object counts between 3
and 30 and at least one relationship. This pre-processing gave us 62,565 training, 5,506 validation, and 5,088 test images with an
average of 10 objects and five relationships per image.
Visual Genome does not provide segmentation masks, so
we omitted the mask prediction loss for models trained on VG.

\subsection{Qualitative Results}
Sample images are shown in Figure \ref{fig:coco} for COCO-stuff, and Figure \ref{fig:visual} for Visual Genome. We observe anecdotally that scene context helps in preserving relationship types and also generate more realistic images. Quantitative metrics for assessing the quality of generated images are limited in efficacy, especially in this task where scene graph compliance is important. We therefore performed subjective evaluations on Mechanical Turk to compare the performance of our model to Johnson \etal \cite{img_gen_from_scene_graph18}. Each query was rated by five independent workers. In approximately 10\% of the trials, we used ground truth images to ensure task compliance and filtered out bad actors.

\paragraph{COCO.}

\begin{figure}

\begin{subfigure}[a]{0.45\textwidth}
\centering
\includegraphics[width=\textwidth]{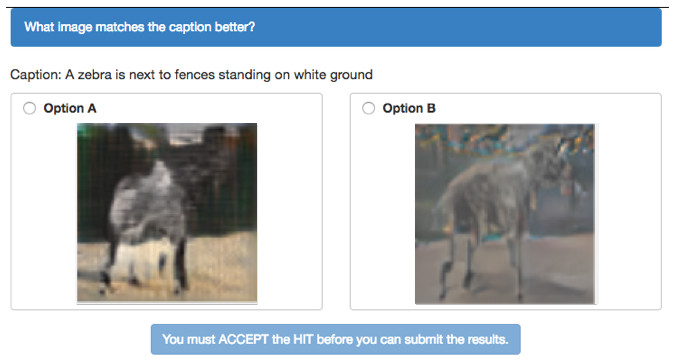}
\end{subfigure}

\begin{subfigure}[b]{0.45\textwidth}
\centering
\begin{tabular}{llll}
- & JJ \cite{img_gen_from_scene_graph18} &  \textbf{Ours}   \\
\hline
 GT box and GT mask &  39.5\% & \textbf{60.5\%}  \\
 Predicted box and mask & \textbf{63.8\%} & 36.2\% \\
 \hline
\end{tabular}

\end{subfigure}
\caption{AB-X comparison inspired by Fig.7 in Johnson \etal \cite{img_gen_from_scene_graph18} on images drawn from the COCO-stuff test set. $N=991$ images in the test set were rated by workers to determine which image matches the caption better. See top for an example query (A = Ours, B = JJ model).}
\label{fig:mscoco_abx}
\end{figure}

For COCO-stuff, we leveraged the included captions to perform an AB-X comparison, inspired by Johnson \etal \cite{img_gen_from_scene_graph18}, where we asked raters in a two-alternative force choice task to select ``which image matches the caption better''. As shown in Figure \ref{fig:mscoco_abx}, when the ground truth box and mask are provided, our model outperforms the Johnson \etal model by a significant margin (60.5\% to 39.5\%). However, our scene graph context model performs worse when generating images using the predicted box and mask. We speculate that the scene graph context and matching loss renders the model less tolerant to noisy box and mask predictions, since the model was never trained under those conditions.

\begin{figure}

\begin{subfigure}[a]{0.45\textwidth}
\centering
\includegraphics[width=\textwidth]{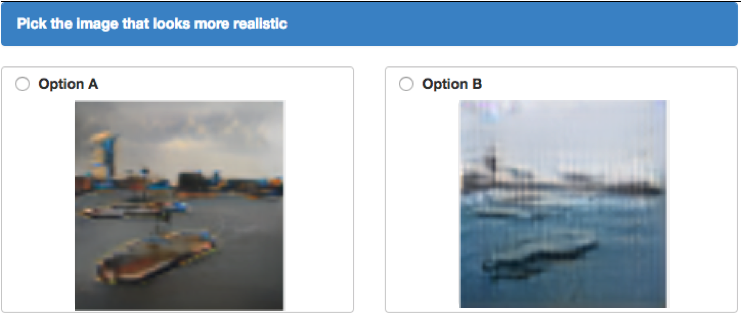}
\end{subfigure}

\begin{subfigure}[b]{0.45\textwidth}
\centering
\begin{tabular}{llll}
- & JJ \cite{img_gen_from_scene_graph18} &  \textbf{Ours}   \\
\hline
 GT box and GT mask & 36.9\% & \textbf{63.1\%}   \\
 Predicted box and mask & \textbf{67.8\%} & 32.2\% \\
 \hline
\end{tabular}

\end{subfigure}
\caption{AvB comparison on images drawn from the COCO-stuff test set. Each worker was asked to rate which image looked more realistic. We tested $N=991$ test set images, with five independent ratings per image.}
\label{fig:mscoco_avb}
\end{figure}

We also carried out an A vs. B test by presenting paired images from the two models and asking workers to select the image that looked more realistic. The results of this user test confirmed the previous findings that our model performed better when provided with ground truth box and masks, but were not as robust to the noisier predicted boxes and masks (Figure \ref{fig:mscoco_avb})

\paragraph{Visual Genome.}

Visual Genome has more complex relationships compared to those derived from COCO-stuff, so we hypothesized that our model's scene context network would provide more of a benefit. For this dataset, we also conducted AB-X and A-B testing against Johnston \etal \cite{img_gen_from_scene_graph18} to measure performance in preserving spatial relationships. For the AB-X test, since captions do not exist in Visual Genome, we randomly sampled relations to generate pseudo-captions (``person on top of grass'') and asked workers to select which image from the two models better matched the relation. 

These user studies revealed that our model outperformed Johnson \etal in terms of both generating realistic images (A-B test, 58\% compared to 42\%), and also generating images that better comply with the scene graph (AB-X, 57\% compared to 43\%). 

For our qualitative studies in this section, we have asked workers to directly compare generated images from both models on image quality. In the next section, we use standalone metrics that measure compliance to the scene graph's spatial layout and semantic relationships.

\begin{table}[]
\centering
\begin{tabular}{llll}
 Study & JJ \cite{img_gen_from_scene_graph18} &  \textbf{Ours}   \\
\hline
  AB-X (Caption) & 42.6\% & \textbf{57.4\%} \\
 \hline
  AvB & 42.7\% & \textbf{58.3\%}\\
 \hline
\end{tabular}
\caption{Visual Genome qualitative study results. Our model outperformed Johnson \etal when workers were asked to select which image was more realistic (AvB), and which image better matched a provided pseudo-caption (AB-X).}
\end{table}


\subsection{Layout Prediction}
\begin{figure}[!t] \begin{center}
\begin{subfigure}[b]{
0.5\textwidth}
    \includegraphics[width=\textwidth]{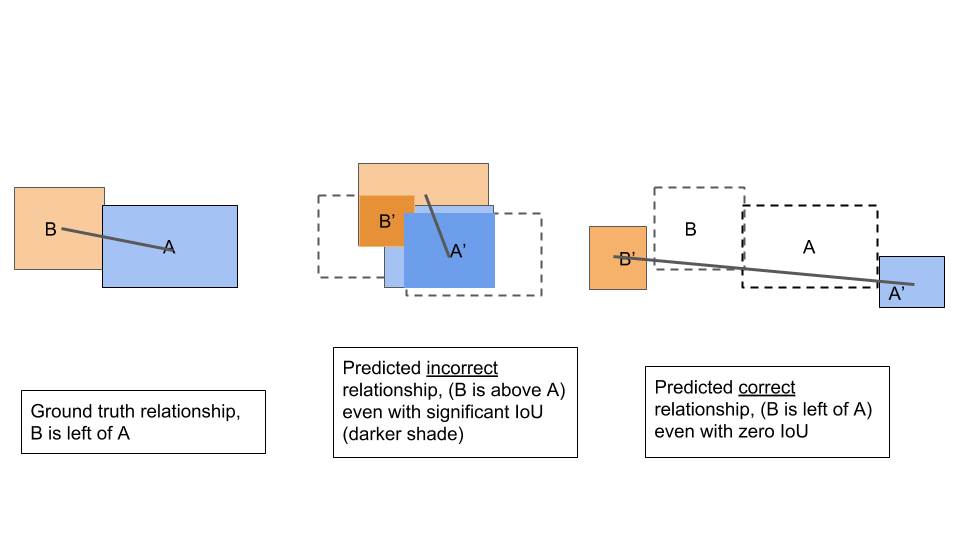}
  \end{subfigure}
\caption{Relation score metric vs IoU. Left: Example ground truth relationship. Center: A high IoU does not guarantee compliance with the intended geometric relationship. Right: Even with $0$ IoU the predicted relationship could still be compliant with the intended relationship }  
\label{fig:relation_score}
\end{center}\end{figure}

In addition to evaluating final image quality, we also compared model performance at the intermediate stage of layout prediction (``Scene layout'' in Figure \ref{fig:pipeline}). Although Intersection-over-Union (IoU) was previously used to measure agreement with the ground truth image, IoU is not the best metric to measure how well geometric relationships between objects in the predicted image comply with the input scene graph. Although IoU may appear as a strict measurement for localization, it is not an indicator of a compliant layout. IoU may not correlate with the geometric relationship as depicted in Figure \ref{fig:relation_score}. The inferred relationship could be incorrect even with significant IoU. Even zero IoU (no overlap with ground truth) could still preserve the relationship indicated in the scene graph.

\paragraph{Relation Score.}
We instead propose a new metric, \emph{relation score}, that measures the compliance of geometric or spatial relationships more accurately than IoU. As an example, if the scene graph specifies that object A is on the left of object B, it is sufficient for the model to comply with that relationship, without requiring the size of the objects to match the ground truth image. See Figure \ref{fig:relation_score} for a graphical illustration.

Because the COCO-stuff relationships are all mutually exclusive and spatial (e.g. 'above', 'below'), we can automate the relation score calculation to verify compliance with the scene graph relationships. We define the relation score as the fraction of spatial relationships that are satisfied in each model's predicted layout. Our scene context model outperforms Johnson \etal in both metrics: IoU (0.483 vs. 0.459) and the relation score (0.54 vs. 0.51), as shown in Table \ref{tab:rel_score_coco}.

\begin{table}[]
\centering
\begin{tabular}{llll}
\textbf{Metric} & \textbf{JJ \cite{img_gen_from_scene_graph18}} &  \textbf{Ours}  \\
\hline
 Avg IOU & 0.459 &  \textbf{0.483} \\
 Relation score & 0.512 & \textbf{0.536} \\
 \hline
\end{tabular}
\caption{Relationship compliance. Relation score (the higher the better) on COCO stuff test set.}
\label{tab:rel_score_coco}
\end{table}

\paragraph{Mean Opinion Relation Score.} The relationship vocabulary in Visual Genome (VG) is rich, and not limited to spatial relationships, so automated computation of relation score metric isn't possible for VG. Instead, we propose a Mean Opinion Relation Score (MORS) metric for relationship compliance. We first analyze the types of relationships contained in this dataset.

\begin{figure}
    \centering
    \includegraphics[width=0.4\textwidth]{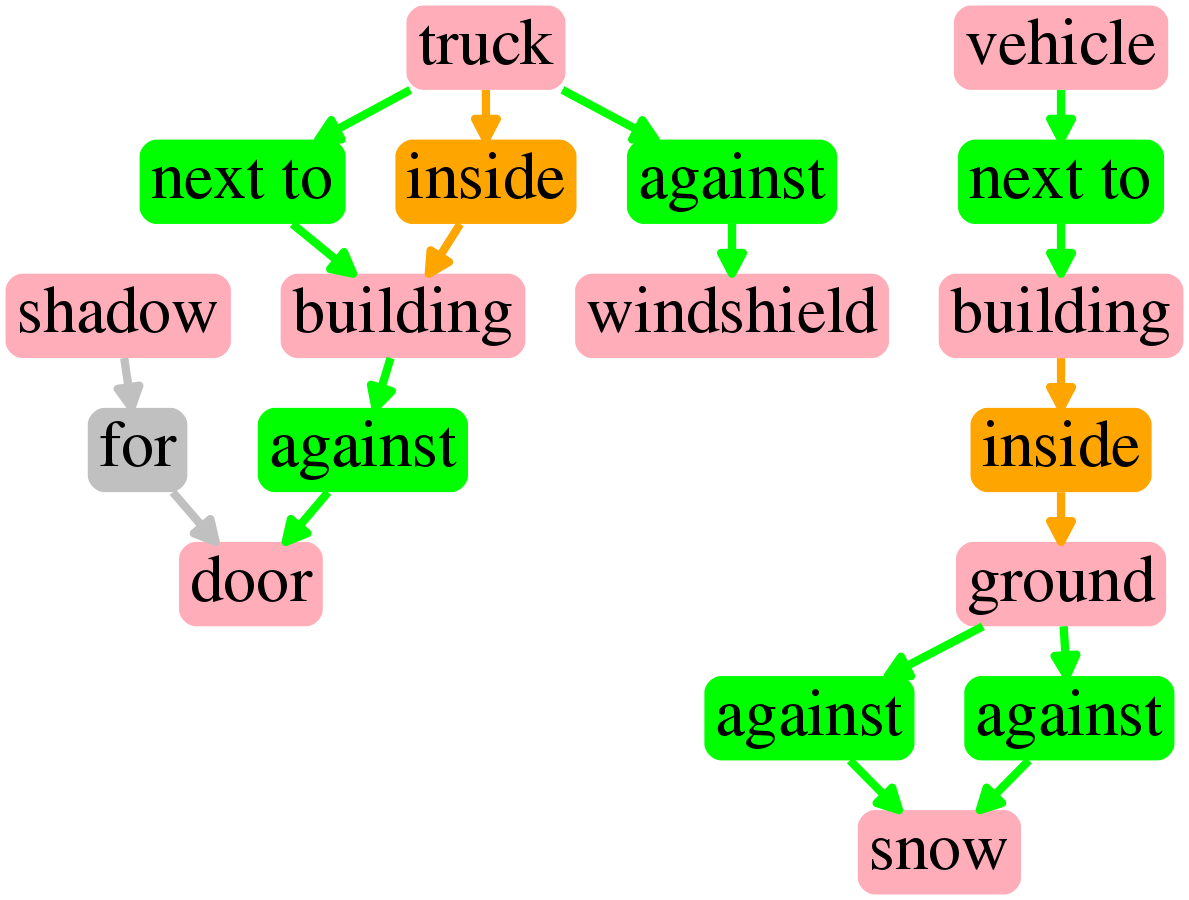}
    \caption{Example scene graph from the Visual Genome dataset, with relationships between objects colored according to their category: Geometric (green), Possessive (orange) and Miscellaneous (grey).}
    \label{fig:color_sg}
\end{figure}

We used the relation types categories from Zellers \etal \cite{neural_motifs_2017}. Of the $46$ relations in Visual Genome, 45\% of the relations are classified as \textit{Geometric}, indicating a spatial relationship. \textit{Semantic} (holding, carrying, walking, etc) and \textit{Possessive} (belonging to, have, etc) constituted 11\% and 10\% of the relationships, respectively. The remainder were marked as \textit{Miscellaneous}, which included descriptors such as: 'and', 'for', or 'of'. This classification is depicted in Table \ref{tab:relclasses}. Figure \ref{fig:color_sg} shows a sample scene graph, with the relationships color-coded (Geometric: green, Semantic: blue, Possessive: orange, and Miscellaneous: grey). 

\begin{figure}

\begin{subfigure}[a]{0.45\textwidth}
\centering
\includegraphics[width=0.7\textwidth]{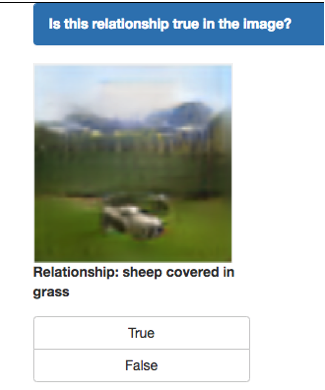}
\end{subfigure}

\begin{subfigure}[b]{0.45\textwidth}
\centering
\begin{tabular}{llll}
\textbf{Relation category} & \textbf{JJ} \cite{img_gen_from_scene_graph18} &  \textbf{Ours} \\
\hline
Semantic & 0.60 & \textbf{0.78} \\
Geometric & 0.64 & \textbf{0.68} \\
Possessive & 0.62 & \textbf{0.80} \\
Miscellaneous & 0.78 & \textbf{0.86} \\
\hline
Overall MORS & 0.64 & \textbf{0.74} & \\
\hline
Avg IOU & 0.223 &  \textbf{0.234}   \\
\end{tabular}

\end{subfigure}
\caption{Mean Opinion Relation score (MORS) on 100 random images and relationship pairs generated from the colored scene graphs in Visual Genome test set. The score is broken by relation category. Each image was rated by five workers. IoU corresponds to all predicted boxes in the test set.}
\label{fig:mors}
\end{figure}

To compute the MORS, we selected single image-relationship pairs, and asked workers to rate whether the relationship is true in the image (Figure \ref{fig:mors}, top). MORS is then defined as the fraction of tested relationships that were found present in the generated image. Visual Genome has several well-known issues, such as semantically overlapping categories, non-exhaustive annotations, and noisy annotations. We therefore manually curated accurate annotations to determine the score. Our results are shown in Figure \ref{fig:mors}. Our model's overall MORS was higher than that of Johnson \etal (0.74 compared to 0.64). We broke down performance by the relation category and observed that for geometric relationships, our models are relatively close (0.68 compared to 0.64), which we expect due to the layout placement mechanism in both models. However, our model was significantly better on non-spatial relationships such as semantic (0.78 vs 0.60) or possessive (0.80 vs 0.62). Our scene context network also includes semantic embeddings, which could be responsible for this improvement.

Note that the relation score measures scene graph compliance at the scene layout stage, with the bounding boxes and segmentation masks. In contrast, MORS can measure more complex non-spatial relationships, and tracks compliance of the final generated image. For more details on the qualitative studies, see the supplemental information.

\begin{table}[]
\centering
\begin{tabular}{ll}
\textbf{Relation category} & \textbf{Relations} \\
\hline
 Semantic & covering, eating, standing on,   \\
  & carrying, looking at, walking on, \\
  & sitting on, sitting in, standing in, \\
  & holding, riding \\
  \hline
 Geometric & next to, above, beside, behind, by, \\
 & laying on, hanging on, under, on,  \\
 & below, against, attached to, near, \\
 & on top of, at, in front of, around, \\
 & along, on side of, parked on\\
 \hline
 Possessive & has, belonging to, inside, with, over, \\
 & covered in, have, in, wears, wearing\\
 \hline
 Miscellaneous & and, for, of, made of\\
 \hline
\end{tabular}
\caption{Relationship categories observed in Visual Genome, inspired by Zeller \etal \cite{neural_motifs_2017}.}
\label{tab:relclasses}
\end{table}

\newcommand{\vcs}{0.18}

\begin{figure*}[!t] \begin{center}
\begin{subfigure}[b]{\vcs\textwidth}
    \includegraphics[width=\textwidth]{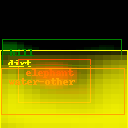}
  \end{subfigure}
  \begin{subfigure}[b]{\vcs\textwidth}
    \includegraphics[width=\textwidth]{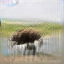}
  \end{subfigure}
  \begin{subfigure}[b]{\vcs\textwidth}
    \includegraphics[width=\textwidth]{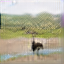}
  \end{subfigure}
\begin{subfigure}[b]{\vcs\textwidth}
    \includegraphics[width=\textwidth]{images/coco/000028_gt_img.png}
  \end{subfigure}
  
  \begin{subfigure}[b]{\vcs\textwidth}
    \includegraphics[width=\textwidth]{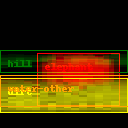}
  \end{subfigure}
  \begin{subfigure}[b]{\vcs\textwidth}
    \includegraphics[width=\textwidth]{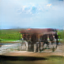}
  \end{subfigure}
  \begin{subfigure}[b]{\vcs\textwidth}
    \includegraphics[width=\textwidth]{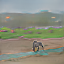}
  \end{subfigure}
\begin{subfigure}[b]{\vcs\textwidth}
    \includegraphics[width=\textwidth]{images/coco/000028_gt_img.png}
  \end{subfigure}
   \vspace{3.00mm}

\begin{subfigure}[b]{\vcs\textwidth}
    \includegraphics[width=\textwidth]{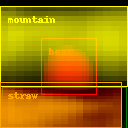}
  \end{subfigure}
  \begin{subfigure}[b]{\vcs\textwidth}
    \includegraphics[width=\textwidth]{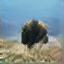}
  \end{subfigure}
  \begin{subfigure}[b]{\vcs\textwidth}
    \includegraphics[width=\textwidth]{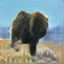}
  \end{subfigure}
\begin{subfigure}[b]{\vcs\textwidth}
    \includegraphics[width=\textwidth]{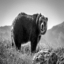}
  \end{subfigure}
  
  \begin{subfigure}[b]{\vcs\textwidth}
    \includegraphics[width=\textwidth]{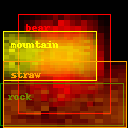}
  \end{subfigure}
  \begin{subfigure}[b]{\vcs\textwidth}
    \includegraphics[width=\textwidth]{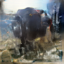}
  \end{subfigure}
  \begin{subfigure}[b]{\vcs\textwidth}
    \includegraphics[width=\textwidth]{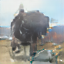}
  \end{subfigure}
\begin{subfigure}[b]{\vcs\textwidth}
    \includegraphics[width=\textwidth]{images/coco/000734_gt_img.png}
  \end{subfigure}
  \vspace{3.00mm}

\begin{subfigure}[b]{\vcs\textwidth}
    \includegraphics[width=\textwidth]{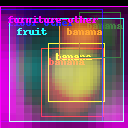}
  \end{subfigure}
  \begin{subfigure}[b]{\vcs\textwidth}
    \includegraphics[width=\textwidth]{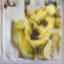}
  \end{subfigure}
  \begin{subfigure}[b]{\vcs\textwidth}
    \includegraphics[width=\textwidth]{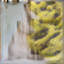}
  \end{subfigure}
\begin{subfigure}[b]{\vcs\textwidth}
    \includegraphics[width=\textwidth]{images/coco/000897_gt_img.png}
  \end{subfigure}
  
  \begin{subfigure}[b]{\vcs\textwidth}
    \includegraphics[width=\textwidth]{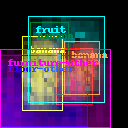}
        \caption{Predicted layout}
  \end{subfigure}
  \begin{subfigure}[b]{\vcs\textwidth}
    \includegraphics[width=\textwidth]{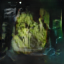}
    \caption{Generated Image}
  \end{subfigure}
  \begin{subfigure}[b]{\vcs\textwidth}
    \includegraphics[width=\textwidth]{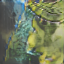}
    \caption{With GT layout}
  \end{subfigure}
\begin{subfigure}[b]{\vcs\textwidth}
    \includegraphics[width=\textwidth]{images/coco/000897_gt_img.png}
    \caption{Ground truth}
  \end{subfigure}
\caption{Examples from COCO-stuff. Sample results of layout prediction, image generation, image generation with ground truth layout for query scene graphs from COCO-stuff test set. For each pair of rows, the top corresponds to our model, and the bottom to results from Johnson \etal \cite{img_gen_from_scene_graph18}. In these anecdotal examples, scene context helps ensure better layout prediction and generation of more realistic images.}  
\label{fig:coco}
\end{center}\end{figure*}



\newcommand{\vgs}{0.18}

\begin{figure*}[!t] \begin{center}
\begin{subfigure}[b]{\vgs\textwidth}
    \includegraphics[width=\textwidth]{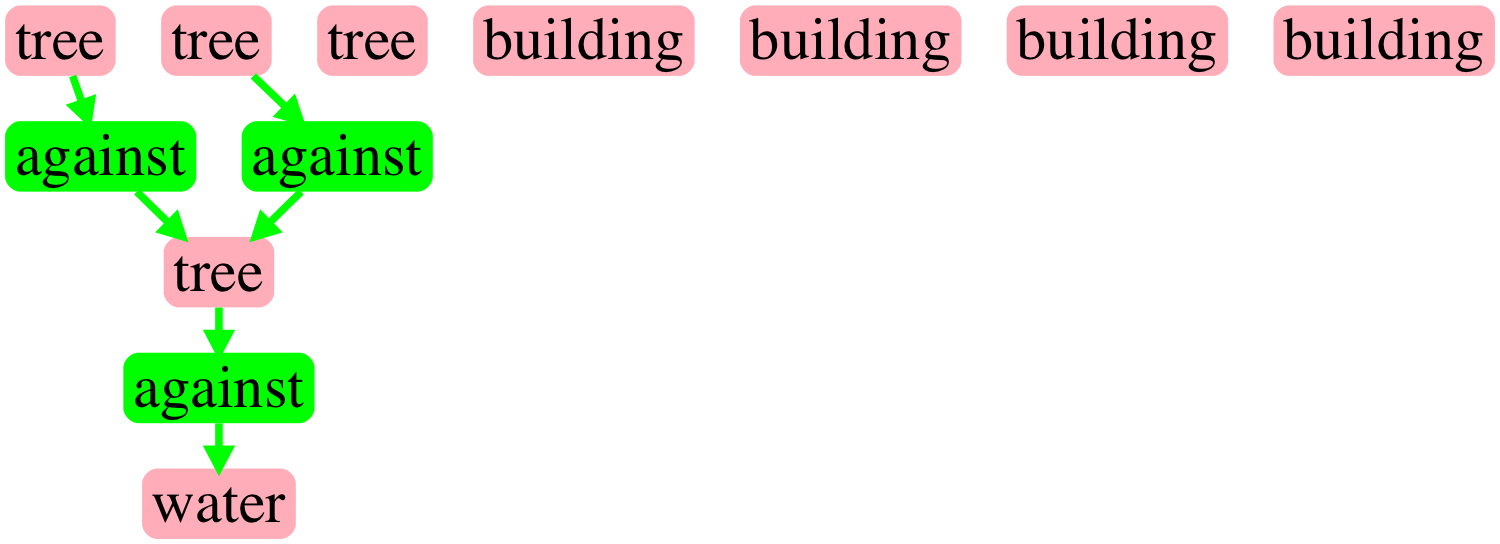}
  \end{subfigure}
  \begin{subfigure}[b]{\vgs\textwidth}
    \includegraphics[width=\textwidth]{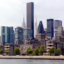}
  \end{subfigure}
  \begin{subfigure}[b]{\vgs\textwidth}
    \includegraphics[width=\textwidth]{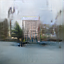}
  \end{subfigure}
\begin{subfigure}[b]{\vgs\textwidth}
    \includegraphics[width=\textwidth]{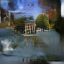}
  \end{subfigure}
   \vspace{3.00mm}


\begin{subfigure}[b]{\vgs\textwidth}
    \includegraphics[width=\textwidth]{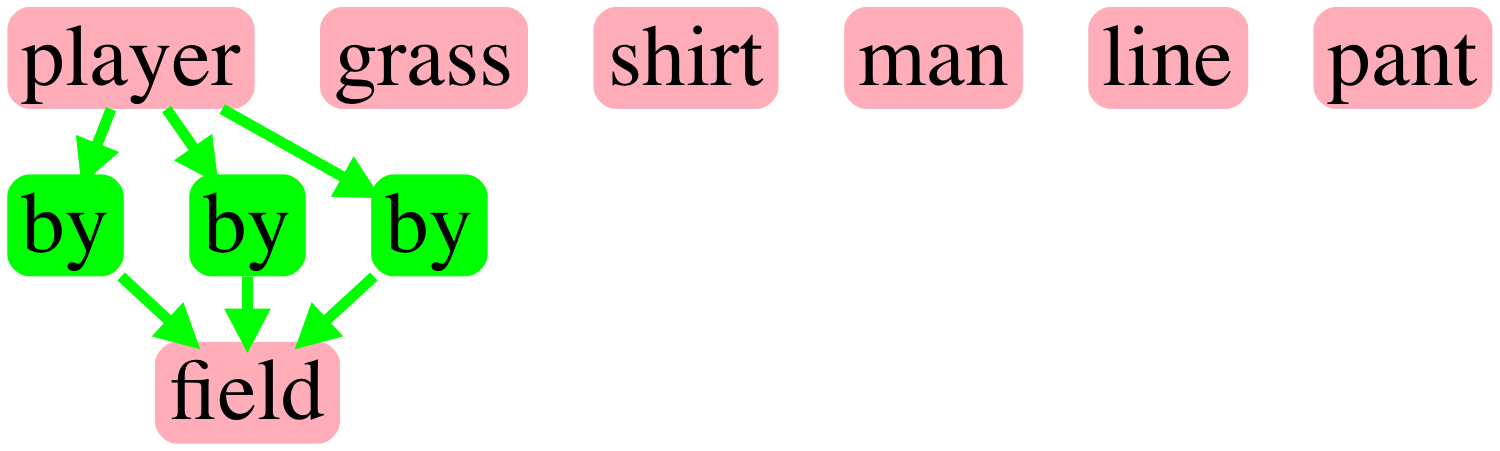}
  \end{subfigure}
  \begin{subfigure}[b]{\vgs\textwidth}
    \includegraphics[width=\textwidth]{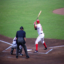}
  \end{subfigure}
  \begin{subfigure}[b]{\vgs\textwidth}
    \includegraphics[width=\textwidth]{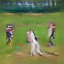}
  \end{subfigure}
  \begin{subfigure}[b]{\vgs\textwidth}
    \includegraphics[width=\textwidth]{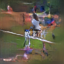}
  \end{subfigure}
   \vspace{3.00mm}
   
    \begin{subfigure}[b]{\vgs\textwidth}
    \includegraphics[width=\textwidth]{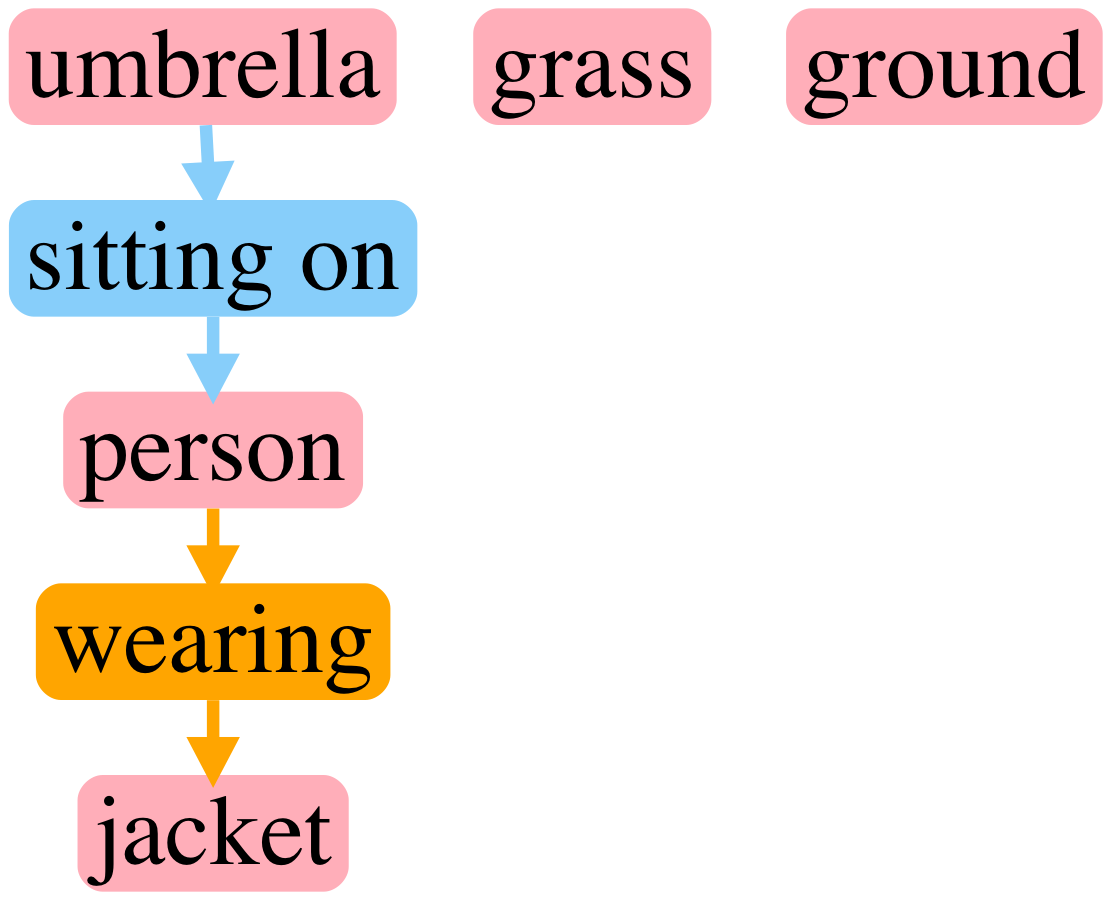}
  \end{subfigure}
  \begin{subfigure}[b]{\vgs\textwidth}
    \includegraphics[width=\textwidth]{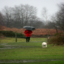}
  \end{subfigure}
  \begin{subfigure}[b]{\vgs\textwidth}
      \includegraphics[width=\textwidth]{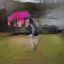}
  \end{subfigure}
  \begin{subfigure}[b]{\vgs\textwidth}
      \includegraphics[width=\textwidth]{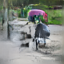}
  \end{subfigure}
   \vspace{3.00mm}
   
   \begin{subfigure}[b]{\vgs\textwidth}
    \includegraphics[width=\textwidth]{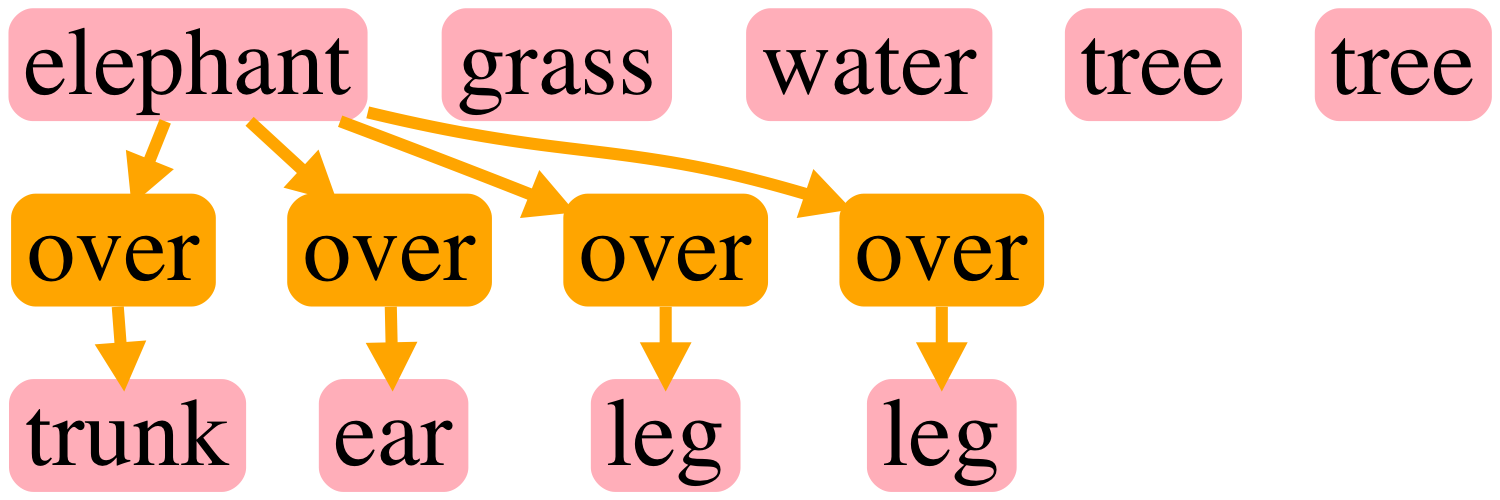}
  \end{subfigure}
  \begin{subfigure}[b]{\vgs\textwidth}
    \includegraphics[width=\textwidth]{images/vg/jj/001020_gt_img.png}
  \end{subfigure}
  \begin{subfigure}[b]{\vgs\textwidth}
    \includegraphics[width=\textwidth]{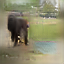}
  \end{subfigure}
  \begin{subfigure}[b]{\vgs\textwidth}
    \includegraphics[width=\textwidth]{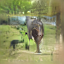}
  \end{subfigure}
   \vspace{3.00mm}
   
   \begin{subfigure}[b]{\vgs\textwidth}
    \includegraphics[width=\textwidth]{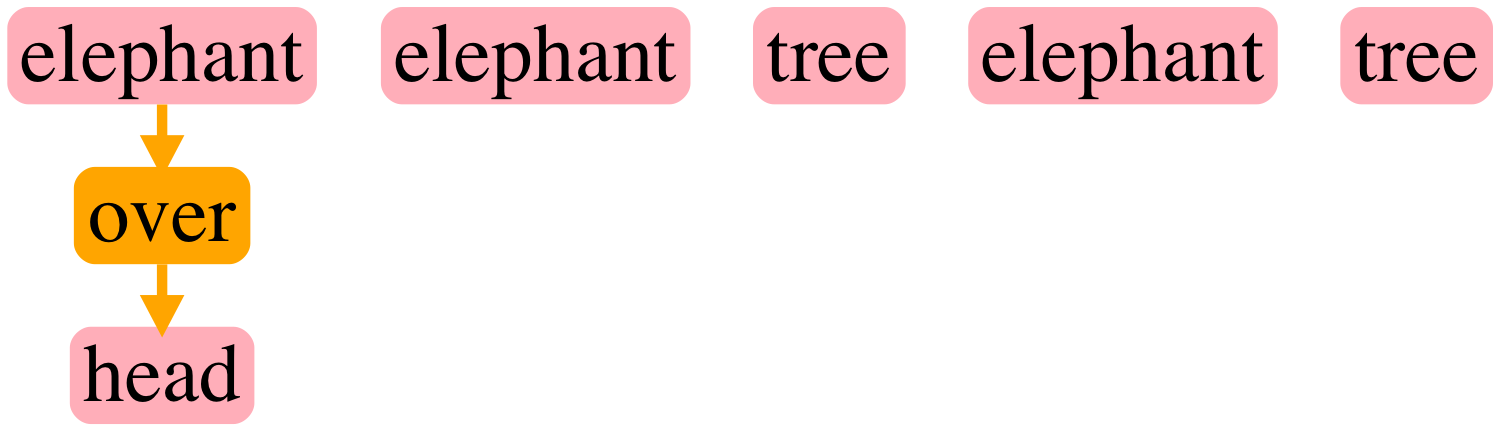}
  \end{subfigure}
  \begin{subfigure}[b]{\vgs\textwidth}
    \includegraphics[width=\textwidth]{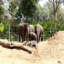}
  \end{subfigure}
  \begin{subfigure}[b]{\vgs\textwidth}
    \includegraphics[width=\textwidth]{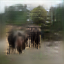}
  \end{subfigure}
  \begin{subfigure}[b]{\vgs\textwidth}
    \includegraphics[width=\textwidth]{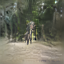}
  \end{subfigure}
   \vspace{3.00mm}
   
   \begin{subfigure}[b]{\vgs\textwidth}
    \includegraphics[width=\textwidth]{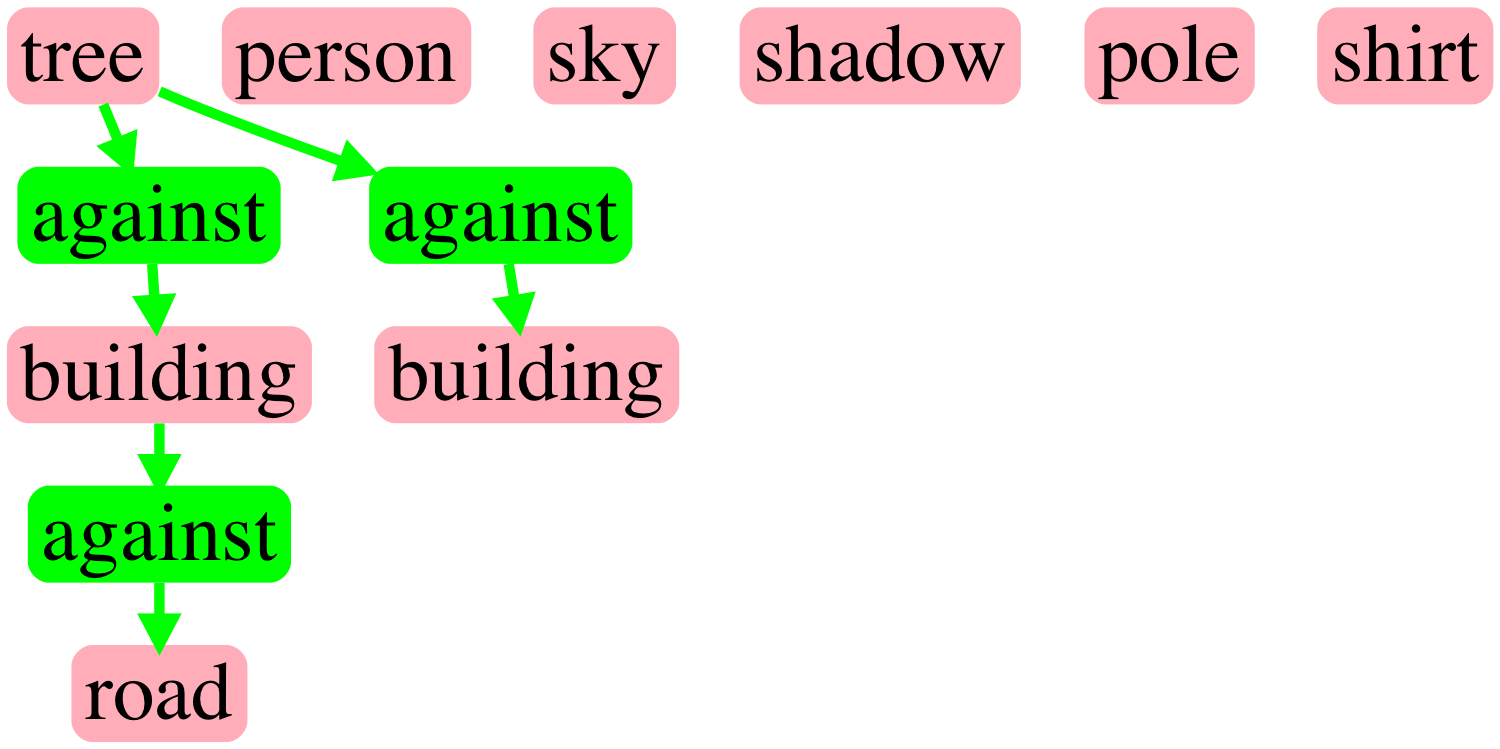}
    \caption{Input Scene Graph}
  \end{subfigure}
  \begin{subfigure}[b]{\vgs\textwidth}
    \includegraphics[width=\textwidth]{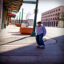}
    \caption{Ground truth}
  \end{subfigure}
  \begin{subfigure}[b]{\vgs\textwidth}
    \includegraphics[width=\textwidth]{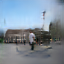}
    \caption{Scene Context (ours)}
  \end{subfigure}
  \begin{subfigure}[b]{\vgs\textwidth}
    \includegraphics[width=\textwidth]{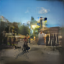}
    \caption{Johnson \etal \cite{img_gen_from_scene_graph18}}
  \end{subfigure}

\caption{Examples from Visual Genome. Generated images on Visual Genome test set. From left to right: (a) input scene graphs, where relationships are color-coded according to relationship types. (b) Ground truth image, and generated images from (c) our model and (d) Johnson \etal \cite{img_gen_from_scene_graph18}. In general, scene context helps in preserving different relationship types among objects.}  
\label{fig:visual}
\end{center}\end{figure*}


\section{Conclusion}
\label{sec:conclusion}
Progress in scene-graph related tasks, such as the image generation task studied here, has been slow for three main reasons. First, datasets such as Visual Genome are hampered by incomplete and incorrect scene graph annotations, or datasets such as COCO-stuff which are synthetic scene graph datasets with relatively simple spatial relationships. Since cleaner data can significantly improve model performance without changing model capacity \cite{siggraph_google_18}, recent efforts have applied heuristic-based methods to better complete annotations \cite{NIPS2017_6628, Paroma_DAWN_blog_2017}. Second, this task lacks metrics designed to directly measure scene graph compliance. Third, new approaches are needed to best integrate scene graph information into the model. 

In this paper, we contributed to two of the three main challenges. We introduced the relation score and mean opinion relation score metrics that measure compliance at the scene layout and generated images stages, respectively. These metrics are more task-relevant than using pixel-based metrics such as intersection-over-union (IoU). Second, we used an auxiliary neural network to encode the scene context for the generator and the discriminator. Conditional  generation on the context yielded images that outperformed the state-of-the-art model \cite{img_gen_from_scene_graph18} on both quantitative metrics such as IoU and relation score, and subjective metrics of scene graph compliance as measured with user studies.

As graph-based tasks increase in number and diversity \cite{battaglia2018relational}, we expect our contributions to generalize to other tasks where either methods to induce graph compliance, or metrics to measure the quality of graph-conditioned output are important. Similarly, future work can borrow from recent progress in generating high-resolution photo-realistic images \cite{karras2017progressive, brock2018large}.

\clearpage

{\small
\bibliographystyle{ieee}
\bibliography{main}

\begin{thebibliography}{10}\itemsep=-1pt

\bibitem{battaglia2018relational}
P.~W. Battaglia, J.~B. Hamrick, V.~Bapst, A.~Sanchez-Gonzalez, V.~Zambaldi,
  M.~Malinowski, A.~Tacchetti, D.~Raposo, A.~Santoro, R.~Faulkner, C.~Gulcehre,
  F.~Song, A.~Ballard, J.~Gilmer, G.~Dahl, A.~Vaswani, K.~Allen, C.~Nash,
  V.~Langston, C.~Dyer, N.~Heess, D.~Wierstra, P.~Kohli, M.~Botvinick,
  O.~Vinyals, Y.~Li, and R.~Pascanu.
\newblock Relational inductive biases, deep learning, and graph networks, 2018.

\bibitem{jnt_embedding17}
E.~Belilovsky, M.~Blaschko, J.~R. Kiros, R.~Urtasun, and R.~Zemel.
\newblock Joint embeddings of scene graphs and images.
\newblock {\em ICLR w/s}, 2017.

\bibitem{brock2018large}
A.~Brock, J.~Donahue, and K.~Simonyan.
\newblock Large scale gan training for high fidelity natural image synthesis,
  2018.

\bibitem{coco-stuff}
H.~Caesar, J.~Uijlings, and V.~Ferrari.
\newblock Coco-stuff: Thing and stuff classes in context.
\newblock In {\em CVPR}, 2018.

\bibitem{crn}
Q.~Chen and V.~Koltun.
\newblock Photographic image synthesis with cascaded refinement networks.
\newblock In {\em CVPR}, 2017.

\bibitem{GIP2018}
R.~Herzig, M.~Raboh, G.~Chechik, J.~Berant, and A.~Globerson.
\newblock Mapping images to scene graphs with permutation-invariant structured
  prediction.
\newblock {\em CoRR}, abs/1802.05451, 2018.

\bibitem{FID_NIPS2017}
M.~Heusel, H.~Ramsauer, T.~Unterthiner, B.~Nessler, and S.~Hochreiter.
\newblock Gans trained by a two time-scale update rule converge to a local nash
  equilibrium.
\newblock In I.~Guyon, U.~V. Luxburg, S.~Bengio, H.~Wallach, R.~Fergus,
  S.~Vishwanathan, and R.~Garnett, editors, {\em Advances in Neural Information
  Processing Systems 30}, pages 6626--6637. Curran Associates, Inc., 2017.

\bibitem{Hong2018CaptionLayout}
S.~Hong, D.~Yang, J.~Choi, and H.~Lee.
\newblock Inferring semantic layout for hierarchical text-to-image synthesis.
\newblock In {\em Computer Vision and Pattern Recognition (CVPR)}, 2018.

\bibitem{img_gen_from_scene_graph18}
J.~Johnson, A.~Gupta, and L.~Fei{-}Fei.
\newblock Image generation from scene graphs.
\newblock {\em CoRR}, abs/1804.01622, 2018.

\bibitem{SG_for_IR_15}
J.~Johnson, R.~Krishna, M.~Stark, L.~Li, D.~A. Shamma, M.~S. Bernstein, and
  L.~Fei-Fei.
\newblock Image retrieval using scene graphs.
\newblock In {\em 2015 IEEE Conference on Computer Vision and Pattern
  Recognition (CVPR)}, volume~00, pages 3668--3678, June 2015.

\bibitem{prog-gan}
T.~Karras, T.~Aila, S.~Laine, and J.~Lehtinen.
\newblock Progressive growing of gans for improved quality, stability, and
  variation.
\newblock In {\em ICLR}, 2018.

\bibitem{karras2017progressive}
T.~Karras, T.~Aila, S.~Laine, and J.~Lehtinen.
\newblock Progressive growing of gans for improved quality, stability, and
  variation.
\newblock In {\em ICLR}, 2018.

\bibitem{adam_KingmaB14}
D.~P. Kingma and J.~Ba.
\newblock Adam: A method for stochastic optimization.
\newblock {\em CoRR}, abs/1412.6980, 2014.

\bibitem{visual_genome16}
R.~Krishna, Y.~Zhu, O.~Groth, J.~Johnson, K.~Hata, J.~Kravitz, S.~Chen,
  Y.~Kalantidis, L.~Li, D.~A. Shamma, M.~S. Bernstein, and F.~Li.
\newblock Visual genome: Connecting language and vision using crowdsourced
  dense image annotations.
\newblock {\em CoRR}, abs/1602.07332, 2016.

\bibitem{coco}
T.-Y. Lin, M.~Maire, S.~Belongie, J.~Hays, P.~Perona, D.~Ramanan, P.~Dollár,
  and C.~L. Zitnick.
\newblock Microsoft coco: Common objects in context.
\newblock In {\em European Conference on Computer Vision (ECCV)}, Zürich,
  2014.
\newblock Oral.

\bibitem{cgan}
M.~Mirza and S.~Osindero.
\newblock Conditional generative adversarial nets.
\newblock {\em CoRR}, abs/1411.1784, 2014.

\bibitem{pixel_to_graph17}
A.~Newell and J.~Deng.
\newblock Pixels to graphs by associative embedding.
\newblock In I.~Guyon, U.~V. Luxburg, S.~Bengio, H.~Wallach, R.~Fergus,
  S.~Vishwanathan, and R.~Garnett, editors, {\em Advances in Neural Information
  Processing Systems 30}, pages 2171--2180. Curran Associates, Inc., 2017.

\bibitem{hourglass_NewellYD16}
A.~Newell, K.~Yang, and J.~Deng.
\newblock Stacked hourglass networks for human pose estimation.
\newblock In {\em Computer Vision - {ECCV} 2016 - 14th European Conference,
  Amsterdam, The Netherlands, October 11-14, 2016, Proceedings, Part {VIII}},
  pages 483--499, 2016.

\bibitem{ma_reed16}
S.~Reed, Z.~Akata, X.~Yan, L.~Logeswaran, B.~Schiele, and H.~Lee.
\newblock Generative adversarial text to image synthesis.
\newblock In M.~F. Balcan and K.~Q. Weinberger, editors, {\em Proceedings of
  The 33rd International Conference on Machine Learning}, volume~48 of {\em
  Proceedings of Machine Learning Research}, pages 1060--1069, New York, New
  York, USA, 20--22 Jun 2016. PMLR.

\bibitem{Reed_text_to_image_2016}
S.~Reed, Z.~Akata, X.~Yan, L.~Logeswaran, B.~Schiele, and H.~Lee.
\newblock Generative adversarial text to image synthesis.
\newblock In {\em Proceedings of the 33rd International Conference on
  International Conference on Machine Learning - Volume 48}, ICML'16, pages
  1060--1069. JMLR.org, 2016.

\bibitem{reed_draw_16}
S.~E. Reed, Z.~Akata, S.~Mohan, S.~Tenka, B.~Schiele, and H.~Lee.
\newblock Learning what and where to draw.
\newblock In D.~D. Lee, M.~Sugiyama, U.~V. Luxburg, I.~Guyon, and R.~Garnett,
  editors, {\em Advances in Neural Information Processing Systems 29}, pages
  217--225. Curran Associates, Inc., 2016.

\bibitem{Paroma_DAWN_blog_2017}
P.~Varma, B.~He, and C.~Ré".
\newblock {\em Exploiting Building Blocks of Data to Efficiently Create
  Training Sets}, 2009 (accessed February 3, 2014).
\newblock \url{https://dawn.cs.stanford.edu//2017/09/14/coral/}.

\bibitem{NIPS2017_6628}
P.~Varma, B.~D. He, P.~Bajaj, N.~Khandwala, I.~Banerjee, D.~Rubin, and
  C.~R\'{e}.
\newblock Inferring generative model structure with static analysis.
\newblock In I.~Guyon, U.~V. Luxburg, S.~Bengio, H.~Wallach, R.~Fergus,
  S.~Vishwanathan, and R.~Garnett, editors, {\em Advances in Neural Information
  Processing Systems 30}, pages 240--250. Curran Associates, Inc., 2017.

\bibitem{siggraph_google_18}
N.~Wadhwa, R.~Garg, D.~E. Jacobs, B.~E. Feldman, N.~Kanazawa, R.~Carroll,
  Y.~Movshovitz-Attias, J.~T. Barron, Y.~Pritch, and M.~Levoy.
\newblock Synthetic depth-of-field with a single-camera mobile phone.
\newblock {\em ACM Trans. Graph.}, 37(4):64:1--64:13, July 2018.

\bibitem{xu2017scenegraph}
D.~Xu, Y.~Zhu, C.~Choy, and L.~Fei-Fei.
\newblock Scene graph generation by iterative message passing.
\newblock In {\em Computer Vision and Pattern Recognition (CVPR)}, 2017.

\bibitem{neural_motifs_2017}
R.~Zellers, M.~Yatskar, S.~Thomson, and Y.~Choi.
\newblock Neural motifs: Scene graph parsing with global context.
\newblock {\em CoRR}, abs/1711.06640, 2017.

\bibitem{stackGan17}
H.~Zhang, T.~Xu, and H.~Li.
\newblock Stackgan: Text to photo-realistic image synthesis with stacked
  generative adversarial networks.
\newblock In {\em {ICCV}}, pages 5908--5916. {IEEE} Computer Society, 2017.

\end{thebibliography}
}

\appendix
\section{Qualitative Studies}
\subsection{Layout Prediction}

Additional examples comparing images generated with our scene context model and those generated from Johnston \etal \cite{img_gen_from_scene_graph18} are shown in Figure \ref{fig:supp_visual}. To highlight the layout, we overlaid the predicted object boundaries over the images. The bounding box and the name of the corresponding object share the same color in the overlaid pictures.

For ease of comparison in this figure, the ground truth images (column d) are flanked by our results (column c) and the Johnston \etal images (column e). In the first row, note the less cluttered scene layout in our model leading to more realistic image generation. In the second row, note that the location of the hands (pink, yellow) and jacket (teal) relative to the head (pink) is more realistic in our model. The third and fourth rows show uncluttered backgrounds (sky and wall), more bounded object shapes, and fewer image generation artifacts. Fifth row shows an example of blurry object (bear) and background (wall and bush) generation. However, our generated objects are more recognizable.

In the last row, both models failed to predict the location of sunglasses (orange), which are both placed far away from the face. However, our scene context model helped in preserving the overall shape of the object from its parts. Overall, we observed that the predicted scene layouts have better compliance with the input scene graph, as well as less cluttered and more realistic relations. This observations on visual inspection of layouts correlate with the Mean Opinion Relation Score (MORS) report in the results section, and reproduced here (Table \ref{tab:mors}). 


\begin{table}[]
\centering
\begin{tabular}{llll}
\textbf{Relation category} & \textbf{JJ} \cite{img_gen_from_scene_graph18} &  \textbf{Ours} \\
\hline
Semantic & 0.60 & \textbf{0.78} \\
Geometric & 0.64 & \textbf{0.68} \\
Possessive & 0.62 & \textbf{0.80} \\
Miscellaneous & 0.78 & \textbf{0.86} \\
\hline
Overall MORS & 0.64 & \textbf{0.74} & \\
\end{tabular}
\caption{Mean Opinion Relation score (MORS) on 100 random images and relationship pairs generated from the colored scene graphs in Visual Genome test set. The score is broken by relation category. Each image was rated by five workers. 
}
\label{tab:mors}
\end{table}

\newcommand{\suppvgs}{0.15}

\begin{figure*}[!t] \begin{center}
  \begin{subfigure}[b]{\suppvgs\textwidth}
    \includegraphics[width=\textwidth]{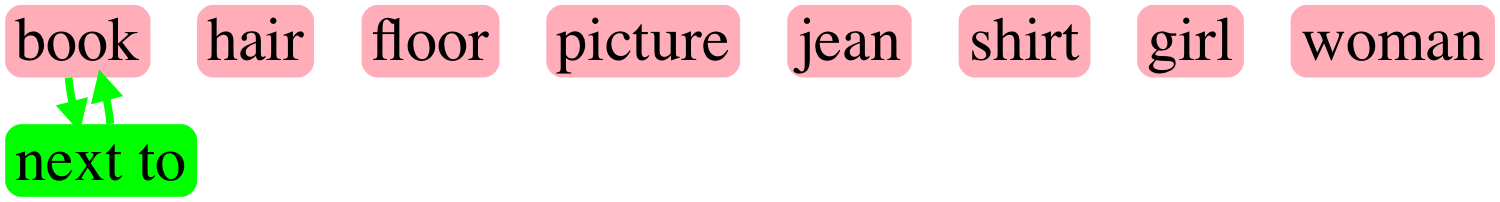}
  \end{subfigure}
\begin{subfigure}[b]{\suppvgs\textwidth}
    \includegraphics[width=\textwidth]{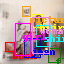}
  \end{subfigure}
  \begin{subfigure}[b]{\suppvgs\textwidth}
    \includegraphics[width=\textwidth]{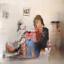}
  \end{subfigure}
  \begin{subfigure}[b]{\suppvgs\textwidth}
    \includegraphics[width=\textwidth]{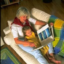}
  \end{subfigure}
\begin{subfigure}[b]{\suppvgs\textwidth}
    \includegraphics[width=\textwidth]{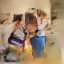}
  \end{subfigure}
\begin{subfigure}[b]{\suppvgs\textwidth}
    \includegraphics[width=\textwidth]{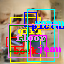}
  \end{subfigure}
   \vspace{3.00mm}
      
\begin{subfigure}[b]{\suppvgs\textwidth}
    \includegraphics[width=\textwidth]{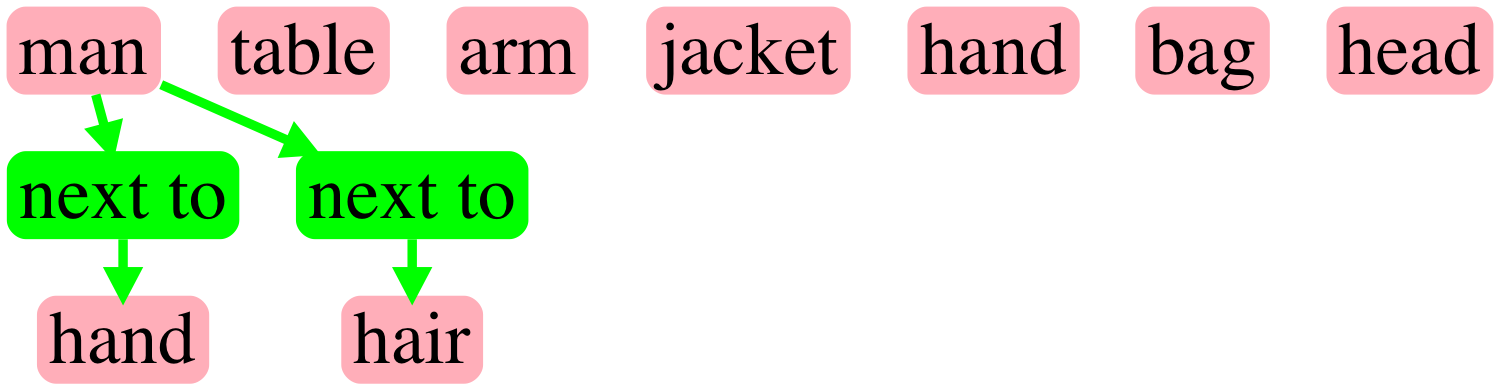}
  \end{subfigure}
\begin{subfigure}[b]{\suppvgs\textwidth}
    \includegraphics[width=\textwidth]{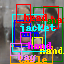}
  \end{subfigure}
  \begin{subfigure}[b]{\suppvgs\textwidth}
    \includegraphics[width=\textwidth]{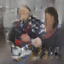}
  \end{subfigure}
  \begin{subfigure}[b]{\suppvgs\textwidth}
    \includegraphics[width=\textwidth]{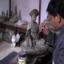}
  \end{subfigure}
\begin{subfigure}[b]{\suppvgs\textwidth}
    \includegraphics[width=\textwidth]{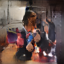}
  \end{subfigure}
\begin{subfigure}[b]{\suppvgs\textwidth}
    \includegraphics[width=\textwidth]{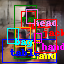}
  \end{subfigure}
   \vspace{3.00mm}

\begin{subfigure}[b]{\suppvgs\textwidth}
    \includegraphics[width=\textwidth]{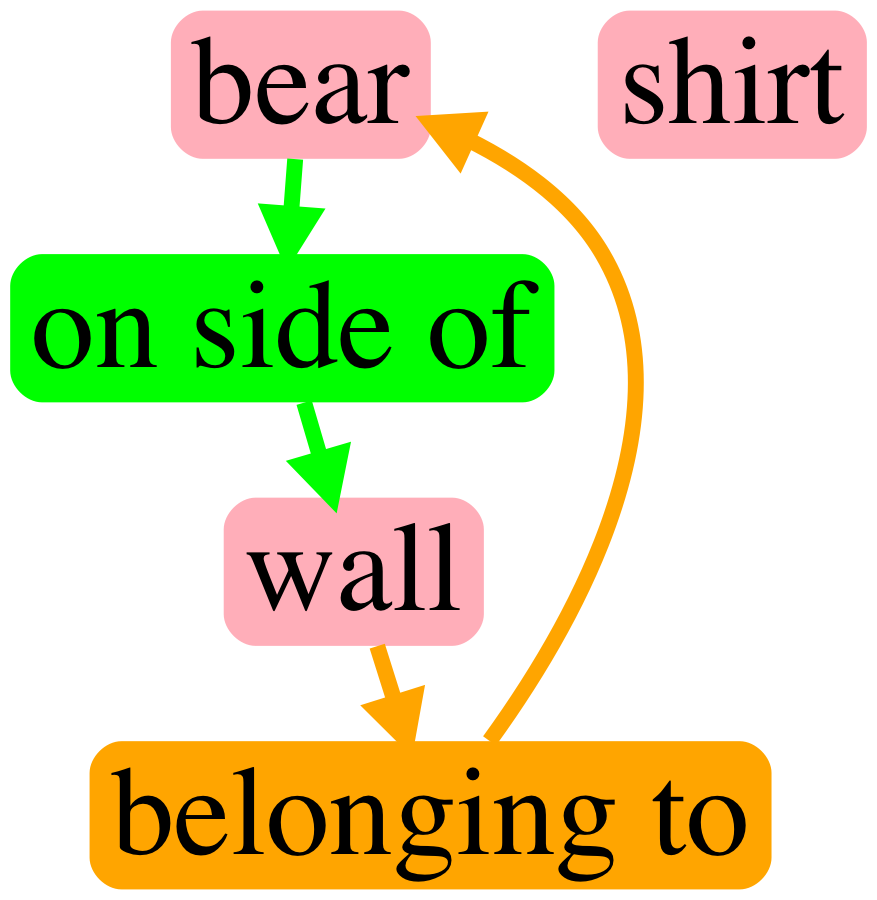}
  \end{subfigure}
\begin{subfigure}[b]{\suppvgs\textwidth}
    \includegraphics[width=\textwidth]{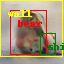}
  \end{subfigure}
  \begin{subfigure}[b]{\suppvgs\textwidth}
    \includegraphics[width=\textwidth]{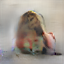}
  \end{subfigure}
  \begin{subfigure}[b]{\suppvgs\textwidth}
    \includegraphics[width=\textwidth]{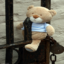}
  \end{subfigure}
\begin{subfigure}[b]{\suppvgs\textwidth}
    \includegraphics[width=\textwidth]{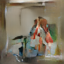}
  \end{subfigure}
\begin{subfigure}[b]{\suppvgs\textwidth}
    \includegraphics[width=\textwidth]{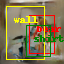}
  \end{subfigure}
   \vspace{3.00mm}   
      
 \begin{subfigure}[b]{\suppvgs\textwidth}
    \includegraphics[width=\textwidth]{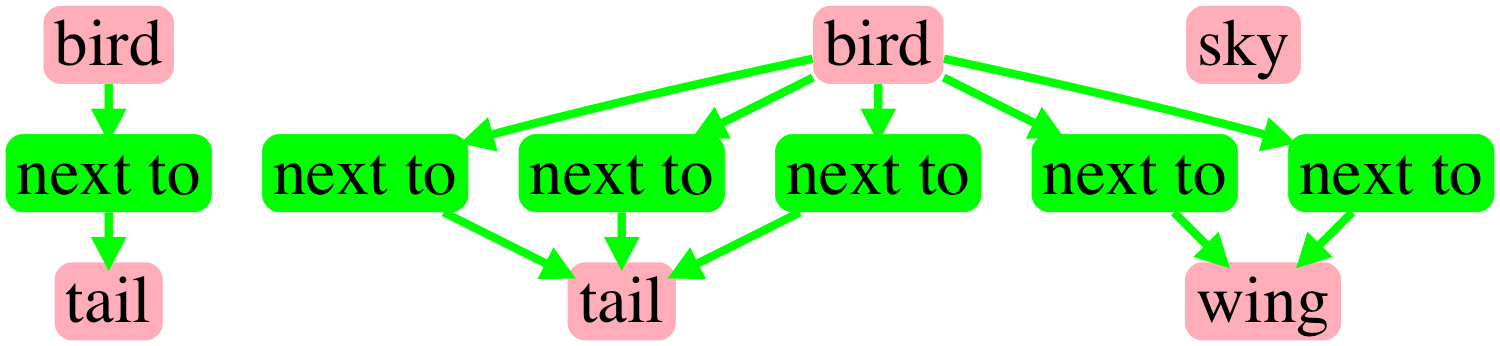}
  \end{subfigure}
\begin{subfigure}[b]{\suppvgs\textwidth}
    \includegraphics[width=\textwidth]{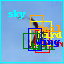}
  \end{subfigure}
  \begin{subfigure}[b]{\suppvgs\textwidth}
    \includegraphics[width=\textwidth]{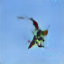}
  \end{subfigure}
  \begin{subfigure}[b]{\suppvgs\textwidth}
    \includegraphics[width=\textwidth]{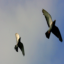}
  \end{subfigure}
\begin{subfigure}[b]{\suppvgs\textwidth}
    \includegraphics[width=\textwidth]{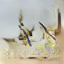}
  \end{subfigure}
\begin{subfigure}[b]{\suppvgs\textwidth}
    \includegraphics[width=\textwidth]{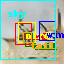}
  \end{subfigure}
   \vspace{3.00mm}

\begin{subfigure}[b]{\suppvgs\textwidth}
    \includegraphics[width=\textwidth]{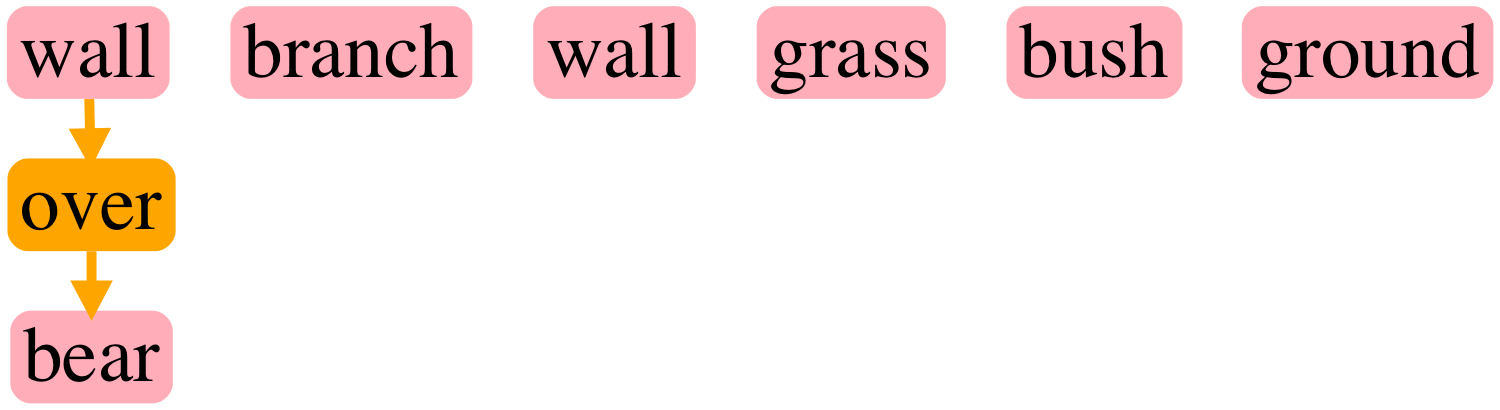}
  \end{subfigure}
\begin{subfigure}[b]{\suppvgs\textwidth}
    \includegraphics[width=\textwidth]{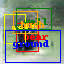}
  \end{subfigure}
  \begin{subfigure}[b]{\suppvgs\textwidth}
    \includegraphics[width=\textwidth]{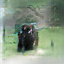}
  \end{subfigure}
  \begin{subfigure}[b]{\suppvgs\textwidth}
    \includegraphics[width=\textwidth]{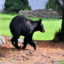}
  \end{subfigure}
\begin{subfigure}[b]{\suppvgs\textwidth}
    \includegraphics[width=\textwidth]{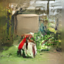}
  \end{subfigure}
\begin{subfigure}[b]{\suppvgs\textwidth}
    \includegraphics[width=\textwidth]{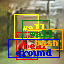}
  \end{subfigure}
   \vspace{3.00mm}  
   
\begin{subfigure}[b]{\suppvgs\textwidth}
    \includegraphics[width=\textwidth]{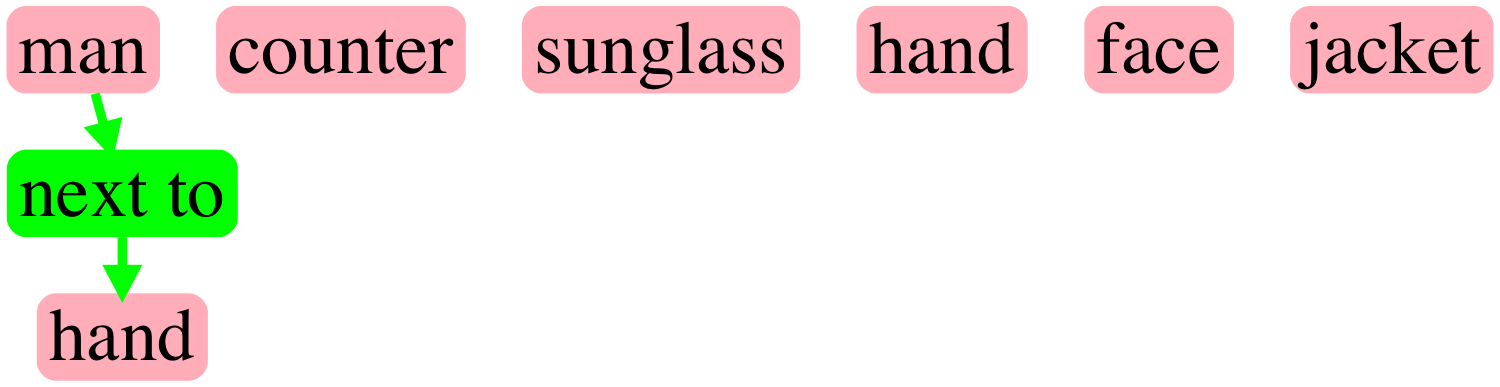}
    \caption{input scene graph}
  \end{subfigure}
\begin{subfigure}[b]{\suppvgs\textwidth}
    \includegraphics[width=\textwidth]{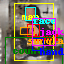}
    \caption{Our layout}
  \end{subfigure}
  \begin{subfigure}[b]{\suppvgs\textwidth}
    \includegraphics[width=\textwidth]{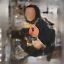}
    \caption{Our result}
  \end{subfigure}
  \begin{subfigure}[b]{\suppvgs\textwidth}
    \includegraphics[width=\textwidth]{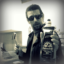}
    \caption{Ground truth}
  \end{subfigure}
\begin{subfigure}[b]{\suppvgs\textwidth}
    \includegraphics[width=\textwidth]{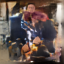}
    \caption{JJ\cite{img_gen_from_scene_graph18} result}
  \end{subfigure}
\begin{subfigure}[b]{\suppvgs\textwidth}
    \includegraphics[width=\textwidth]{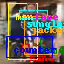}
    \caption{JJ\cite{img_gen_from_scene_graph18} Layout}
  \end{subfigure}

\caption{Examples from Visual Genome. Generated images on Visual Genome test set. From left to right: (a) input scene graphs, where relationships are color-coded according to relationship types. Generated image from our model(c) with overlaid object boundaries (b). (d) Ground truth image, and generated images from (d) Johnson \etal \cite{img_gen_from_scene_graph18} with (e) overlaid object boxes. In general, scene context helps in preserving different relationship types among objects. Best viewed in color.}  
\label{fig:supp_visual}
\end{center}\end{figure*}

\subsection{Image Quality}

In addition to the RGB quality evaluation by workers, we computed the Frechet Inception Distance (FID) 
\cite{FID_NIPS2017} on the entire test set.  Table \ref{tab:fid} shows the FID scores for images generated on COCO-stuff and Visual Genome test set respectively. Lower FID value denotes better image quality and diversity. The Johnston \etal model had better FID scores that ours, even though in the Mechanical Turk experiments reported in our main text, our model produces more realistic images in a direct A versus B comparison. Since the FID is based on feature extraction, we speculate that the FID score is not capturing the spatial relationships between objects in these complex scenes. We note a similar result in Johnston \etal, where their Inception scores were worse than the StackGAN model, even though their images were rated higher. Together, these results suggest that a new quantitative metric is needed for this particular task.

We also observed anecdotally that the color images from our model were less vibrant. We speculate that the \emph{sum pooling} in our scene context network may be introducing the undesirable low contrast effect in the generated images. We attribute the low contrast as one of the primary reasons why our model's output did not have better FID score than the Johnson \etal model. Further investigation on better pooling mechanisms to reduce the artifact remains as a future work.

\begin{table}[tb]
\centering
\begin{tabular}{llll}
FID & JJ \cite{img_gen_from_scene_graph18} &  \textbf{Ours} \\
\hline
COCO-stuff & 92.6\% & 99.6\%  \\
 \hline
  Visual Genome & 94.4\% & 100.7\%  \\
 \hline
\end{tabular}
\caption{FID score on COCO-stuff \cite{coco-stuff} and Visual Genome \cite{visual_genome16} test dataset. Lower FID score generally denotes better quality.}
\label{tab:fid}
\end{table}

\section{Experimental Details}

We carried out several experiments on Mechanical Turk, as described in the main paper. In this section, we provide additional details on the experimental procedure and results. For each study, we selected a subset of scene graphs from the test set, and obtained generated images from both the Johnson \etal paper, as well as our model. We then ran two experiments for COCO-stuff and Visual Genome datasets with a two-alternative forced choice task:

\begin{enumerate}
    \item \textbf{AvB}. Workers were asked which image was more realistic.
    \item \textbf{AB-X (Caption)}. Workers were asked which image better matches the provided caption.
\end{enumerate}

In our experiments, we also inserted the ground truth images in a subset of trials as a positive control. The trial types were randomly interleaved. Table \ref{table:details} shows more details on the number of trials, and accuracy rate on the control trials. Note that, since Visual Genome did not have ground truth masks 
, we only used the predicted mask and bounding boxes in our experiments.

The AB-X comparisons require providing a caption. For COCO, we used the annotated captions. For Visual Genome, because captions did not exist, we manually selected $200$ relationships that were then converted to sentences. The Visual Genome annotations are noisy, so we had to filter relationships that accurately described the images. Due to budgetary constraints, in the COCO AB-X experiments, we randomly sub-sampled $300$ of the $991$ images in the test set.

Compared to the previous experiments, 
the Mean Opinion Relation Score (MORS) 
experiment did not ask workers to decide between two images. Instead, a single image and a corresponding relation was presented, and workers were asked if the stated relationship is true in the image. The images from both models, as well as some ground truth images, were randomly interleaved.

\begin{table*}[tb]
\begin{tabular}{lllrrr}
\hline
Experiment & Dataset       & Image Type         & \# Images & Total \# Trials & \% Correct (Control Trials)\\
\hline
AvB        & COCO          & GT Mask, GT BB     & 991       & 4960                          & 97.2\%                     \\
AvB        & COCO          & Pred Mask, Pred BB & 300       & 1380                         & 91.6\%                     \\
AB-X       & COCO          & GT Mask, GT BB     & 991       & 3862                        & 94.5\%                     \\
AB-X       & COCO          & Pred Mask, Pred BB & 300       & 1380                          & 93.2\%                     \\
AvB        & Visual Genome & Pred Mask, Pred BB & 1018      & 2964                         & 83.3\%                     \\
AB-X       & Visual Genome & Pred Mask, Pred BB & 200       & 500                       & 83.6\%                     \\
MORS & Visual Genome & Pred Mask, Pred BB & 200       & 1000                        & 86.4\%           

\end{tabular}
\caption{Experimental details.}
\label{table:details}
\end{table*}

\end{document}